%% file: main.tex
\documentclass[11pt,a4paper]{article}
\usepackage[hyperref]{naaclhlt2018}
\usepackage{times}
\usepackage{latexsym}
\usepackage[utf8]{inputenc}
\usepackage{todonotes}
\usepackage{amsmath}
\usepackage{breqn}
\usepackage{url}
\usepackage{booktabs}
\usepackage{float}
\aclfinalcopy % Uncomment this line for the final submission

\title{On the difficulty of a distributional semantics of spoken language}

\author{Grzegorz Chrupała \\
  Tilburg University\\
  {\tt g.chrupala@uvt.nl} \\\And
  Lieke Gelderloos\\
  Tilburg University\\
  {\tt l.j.gelderloos@uvt.nl } \\\AND
  Ákos Kádár \\
  Tilburg University\\
  {\tt a.kadar@uvt.nl }\\\And
  Afra Alishahi \\
  Tilburg University\\
  {\tt a.alishahi@uvt.nl}}

\date{}
\hypersetup{draft}
\begin{document}

\maketitle
\begin{abstract}
\input{abstract.tex}
\end{abstract}
%\noindent\textbf{Index Terms}: speech recognition, distributional semantics, unsupervised learning, representation learning

% WRITIING:

%Lieke: Related work
%Afra: Introduction
%Grzegorz: Rest

\section{Introduction}
In the realm of NLP for written language, unsupervised approaches to
inducing semantic representations of words have a long pedigree and a
history of substantial success 
\cite{landauer1998introduction,blei2003latent,mikolov2013distributed}.
The core idea behind these models is to build word representations that
can predict their surrounding context. In search for similarly generic 
and versatile representations of whole sentences, various composition 
operators have been applied on word representations \citep[e.g.\ ][]{socher2013recursive,kalchbrenner2014convolutional,kim2014convolutional,zhao2015self}. Alternatively, sentence representations are induced via the objective to predict the surrounding sentences  
\citep[e.g.\ ][]{le2014distributed, kiros2015skip,arora2016simple,jernite2017discourse,logeswaran2018efficient}. %subramanian2018learning 
Such representations capture aspects of the meaning of the 
encoded sentences, which can be used in a variety of tasks such as semantic entailment 
or text understanding.

In the case of spoken language, unsupervised methods usually focus on 
discovering relatively low-level constructs such as phoneme inventories or 
word-like units. This is mainly due to the fact that 
the key insight from distributional semantics that ``you shall know the word 
by the company it keeps'' \citep{firth1957synopsis} is hopelessly confounded in the case of spoken 
language. In text two words are considered semantically similar 
if they co-occur with similar neighbors. However, speech segments which occur 
in the same utterance or situation often have many other features
in addition to similar meaning, such as being uttered by the same speaker
or accompanied by similar ambient noise. 

In this study we show that if we can 
abstract away from speaker and background noise, we can effectively capture 
semantic characteristics of spoken utterances in an unsupervised way. We present SegMatch, a model trained to match segments of the same utterance. SegMatch utterance encodings are compared to those in Audio2Vec, which is trained to decode the context that surrounds an utterance.
%these last sentences are new - please check if you agree. should we add a sentence describing why we think segmatch is better for semantics?
% GC: I don't think we have a very convincing explanation about why it's better
To 
investigate whether our representations capture semantics, we evaluate on  
speech and vision datasets where photographic images are paired with spoken 
descriptions. Our experiments show that for a single synthetic voice, a simple model 
%\todo{we never talk about any model up to this point. - commented out because we do now}
trained only on image captions can capture pairwise similarities that correlate with those in the visual space.

% In order to facilitate evaluation of the semantic aspect of the learned representations, 
% we work with speech and vision datasets, which couple photographic images with their spoken descriptions. 
% We show that for a single synthetic voice, it is possible to learn meaningful representations of spoken utterances without supervision from the visual modality, and that pairwise similarities of these representations correlate with the pairwise similarities in the visual space. 
Furthermore we discuss the factors preventing effective learning in datasets with multiple human speakers: these include confounds between semantic and situational factors as well as artifacts in the datasets.

\section{Related work}
Studies of unsupervised learning from speech typically aim to discover the phonemic or lexical building blocks of the language signal. \citet{park2008unsupervised} show that words and phrase units in continuous speech can be discovered using algorithms based on dynamic time warping.  \citet{DBLP:journals/corr/abs-1711-00937} introduce a Vector Quantised-Variational AutoEncoder model, in which a convolutional encoder trained on raw audio data gives discrete encodings that are closely related to phonemes.
Recently several unsupervised speech recognition methods were proposed that segment speech and cluster the resulting word-like segments \citep{kamper2017segmental} or encode them into segment embeddings containing phonetic information \citep{wang2018segmental}. \citet{scharenborg2018linguistic} show that word and phrase units arise as a by-product in end-to-end tasks such as speech-to-speech translation.
In the current work, the aim is to directly extract semantic, rather than word form information from speech.

Semantic information encoded in speech is used in studies that ground speech to the visual context. Datasets of images paired with spoken captions can be used to train multimodal models that extract visually salient semantic information from speech, without access to textual information \cite{harwath2015deep, harwath2016unsupervised, kamper2017visually, chrupala2017representations,K17-1037,harwath2017learning}. 
This form of semantic supervision, through contextual information from another modality, has its limits: it can only help to learn to understand speech describing the here and now. 

On the other hand, the success of word embeddings derived by distributional semantic principles has shown how rich the semantic information within the structure of language itself is. 
Semantic representations of words obtained through Latent Semantic Analysis have proven to closely resemble human semantic knowledge \cite{blei2003latent, landauer1998introduction}. Word2vec models produce semantically rich word embeddings by learning to predict the surrounding words in text \cite{mikolov2013efficient,mikolov2013distributed} and this principle is extended to sentences in the Skip-thought model \cite{kiros2015skip} and several subsequent works  \citep{arora2016simple,jernite2017discourse,logeswaran2018efficient}.

In the realm of spoken language, in \citet{chung2017skipgram} the sequence-to-sequence Audio2vec model learns semantic embeddings for audio segments corresponding to words, by predicting the audio segments around it. \citet{chung2018speech2vec} further experiment with this model and rename it to Speech2vec. \citet{chen2018towards} train semantic word embeddings from word-segmented speech as part of their method of training an ASR system from non-aligned speech and text. These works are closely related to our current study, but crucially, unlike them we do {\it not} assume that speech is already segmented into discrete words.

\section{Models}
\label{sec:models}

\subsection{Encoder}
\label{sec:encoder}
All the models in this section use the same encoder architecture. The encoder is loosely based on the architecture 
of \citet{chrupala2017representations}, i.e.\ it consists of a 1-dimensional convolutional layer which subsamples the input, followed by a stack of recurrent layers, followed by a self-attention operator. Unlike \citet{chrupala2017representations} we use GRU layers \cite{chung2014empirical} instead of RHN layers \cite{pmlr-v70-zilly17a}, and do not implement residual connections. These modifications are made in order to exploit the fast native CUDNN implementation of a GRU stack and thus speed up experimentation in this exploratory stage of our research. 
The encoder $\mathrm{Enc}$ is defined as follows:
\begin{equation}
  \label{eq:encode_u}
  \mathrm{Enc}(\mathbf{x}) = \mathrm{unit(\mathrm{Attn}(\mathrm{GRU}_{\ell} (\mathrm{Conv}_{s,d,z}(\mathbf{x}))))}
\end{equation}
where  $\mathrm{Conv}$ is a convolutional layer with length
$s$, $d$ channels, and stride $z$, $\mathrm{GRU}_\ell$ is a stack of $\ell$ GRU layers, $\mathrm{Attn}$ is self-attention and $\mathrm{unit}$ is L2-normalization.
The self-attention operator computes a weighted sum of the RNN activations at all timesteps:
\begin{equation}
  \label{eq:attn}
  \mathrm{Attn}(\mathbf{x}) = \sum_t \alpha_t \mathbf{x}_t
\end{equation}
where the weights $\alpha_t$ are determined by an MLP with learned parameters
$\mathbf{U}$ and $\mathbf{W}$, and passed through the timewise softmax function:
\begin{equation}
  \label{eq:alpha_t}
  \alpha_t = \frac{\exp(\mathbf{U}\tanh(\mathbf{Wx}_t))}{\sum_{t'} \exp(\mathbf{U}\tanh(\mathbf{Wx}_{t'}))}
\end{equation}

\subsection{Audio2vec} 
Firstly we define a model inspired by \citet{chung2017skipgram} which uses the multilayer GRU encoder described above, and a single-layer GRU decoder, conditioned on the output of the encoder. 

The model of  \citet{chung2017skipgram} works on word-segmented speech: the encoder encodes the middle word of a five word sequence, and the decoder decodes each of the surrounding words. Similarly, the Skip-thought model of \cite{kiros2015skip} works with  a sequence of three sentences, encoding the middle one and decoding the previous and next one. In our fully unsupervised setup we do not have access to word segmentation, and thus our Audio2vec models work with arbitrary speech segments. We split each utterance into three equal sized chunks: the model encodes the middle one, and decodes the first and third one.

The decoder predicts the MFCC features at time $t+1$ based on the state of the hidden layer at time $t$. From reading \citet{chung2017skipgram} it is not clear whether in addition to the hidden state their decoder also receives the MFCC frame at $t$ as input. We thus implemented two versions, one with and one without this input.

\vskip0.1cm \noindent{\bf Audio2vec-C~} The decoder receives the output of the encoder as the initial state of the hidden layer, and the frame at $t$ as input as it predicts the next frame at $t+1$.
\begin{align}
\mathbf{\hat{x}}^{\text{first}}_{t+1} & = \mathbf{F}\mathbf{h}_t \\
\mathbf{h}_t     & = \mathrm{gru}(\mathbf{h}_{t-1}, \mathbf{x}^{\text{first}}_{t}) \\
\mathbf{h}_0     & = \mathrm{Enc}\left(\mathbf{x}^{\text{middle}}\right)
\end{align}
where $\mathbf{x}^{\text{first}}_{t}$ are the MFCC features of the previous chunk at time $t$, $\mathbf{\hat{x}}^{\text{first}}_{t+1}$ are the predicted features at the next time step, $\mathbf{F}$ is a learned projection matrix, $\mathrm{gru}(\cdot, \cdot)$ is a single step of the GRU recurrence, and $\mathbf{x}^{\text{middle}}$ is the sequence of the MFCC features of the input.
The decoder for the third chunk $\mathbf{x}^{\text{third}}$ is defined in the same way.

\vskip0.1cm \noindent{\bf Audio2vec-U~} The decoder receives the output of the encoder as the input at each time step, but does not have access to the frame at $t$.
\begin{align}
\mathbf{\hat{x}}^{\text{first}}_{t+1} & = \mathbf{F}\mathbf{h}_t \\
\mathbf{h}_t     & = \mathrm{gru}(\mathbf{h}_{t-1}, \mathrm{Enc}(\mathbf{x}^{\text{middle}}))
\end{align}
In this version $\mathbf{h}_0$ is a learned parameter.
There are two separate decoders: i.e.\ the weights of the decoder for the first chunk and for the third chunk are {\em not} shared.

For both versions of {\bf Audio2vec} the loss function is the Mean Squared Error. 

\subsection{SegMatch}
This model works with segments of utterances also: we split each utterance approximately in half, while erasing a short portion in the center in order to prevent the model from finding trivial solutions based on matching local patterns at the edges of the segments.
The encoder is as described above. After encoding the segments, we project the initial and final segments via separate learned projection matrices:
\begin{align}
\mathbf{b} & =  \mathbf{B} \mathrm{Enc}(\mathbf{x}_{0:m}) \\
\mathbf{e} & =  \mathbf{E} \mathrm{Enc}(\mathbf{x}_{m+k:n}) 
\end{align}
where $\mathbf{x}_{0:n}$ is the sequence of MFCC frames for an utterance, $k$ is the size of the erased segment, $\mathrm{Enc}(\cdot)$ is the encoder and $\mathbf{B}$ and $\mathbf{E}$ are the projection matrices for the beginning and end segment respectively. That is, there is a single shared encoder for both types of speech segments (beginning and end), but the projections are separate.
There is no decoding, but rather the model learns to match encoded segments from the same utterance and distinguish them from encoded segments from different utterances within the same minibatch. The loss function is similar to the one for matching spoken utterances to images in \citet{chrupala2017representations}, with the difference that here we are matching utterance segments to each other:

\begin{dmath}
\mathcal{L} =  \sum_{\mathbf{b},\mathbf{e}} \left(\sum_{\mathbf{b}'} \max[0, \alpha + d(\mathbf{b},\mathbf{e}) - d(\mathbf{b}',\mathbf{e})] +
    \sum_{\mathbf{e}'} \max[0, \alpha + d(\mathbf{b},\mathbf{e}) - d(\mathbf{b},\mathbf{e}')] \right)
\end{dmath}
where $(\mathbf{b,e})$ are beginning and end segments from the same utterance, and $(\mathbf{b',e})$ and $(\mathbf{b,e'})$ are beginning and end segments from two different utterances within a batch, while $d(\cdot, \cdot)$ is the cosine distance between encoded segments. The loss function thus attempts to make the cosine distance between encodings of matching segments less than the distance
between encodings of mismatching segment pairs, by a margin.

\vskip 0.2cm
\noindent Note that the specific way we segment speech is not a crucial component of either of the models: it is mostly driven by the fact that we run our experiments on speech and vision datasets, where speech consists of isolated utterances. For data consisting of longer narratives, or dialogs, we could use different segmentation schemes.

\section{Experimental  setup}

\subsection{Datasets}
In order to facilitate evaluation of the semantic aspect of the learned representations, 
we work with speech and vision datasets, which couple photographic images with their spoken descriptions. Thanks to the structure of these data we can use the evaluation metrics detailed in section~\ref{sec:metrics}.

%The image
%features come from the final fully connect layer of VGG-16
%\citep{simonyan2014very} pre-trained on Imagenet \cite{ILSVRCarxiv14}.

\vskip0.1cm \noindent{\bf Synthetically spoken COCO~}
This dataset was created by \citet{chrupala2017representations}, based on the original COCO dataset \cite{lin2014microsoft}, using the Google TTS API. The captions are spoken by a single synthetic voice, which is realistic but simpler than human speakers, lacking variability and ambient noise. There are 300,000 images, each with five captions. Five thousand images each are held out for validation and test.

\vskip0.1cm \noindent{\bf Flickr8k Audio Caption Corpus~}
This dataset \cite{harwath2015deep} contains the captions in the original Flickr8K
corpus \cite{hodosh2013framing} read aloud by crowdworkers. There are 
8,000 images, each image with five descriptions. One thousand images
are held out for validation, and another one thousand for the  test set. %We use the splits provided by \citep{karpathy2015deep}. 

\vskip0.1cm \noindent{\bf Places Audio Caption Corpus~}
This dataset was collected by \cite{harwath2016unsupervised} using crowdworkers. Here each image is described by a single spontaneously spoken caption. There are 214,585 training images, and 1000 validation images (there are no separate test data).

\subsection{Evaluation metrics} 
\label{sec:metrics}
We evaluate the quality of the learned semantic speech representations according to the following criteria. 

\vskip0.1cm \noindent{\bf Paraphrase retrieval~} For the Synthetically Spoken COCO dataset as well as for the Flickr8k Audio Caption Corpus each image is described via five independent spoken captions. Thus captions describing the same image are effectively paraphrases of each other. This structure of the data allows us to use a paraphrasing retrieval task as a measure of the semantic quality of the learned speech embeddings. 
We encode each of the spoken utterances in the validation data, and rank the others according to the cosine similarity. We then measure: (a) Median rank of the top-ranked paraphrase; and (b) recall@K: the proportion of paraphrases among $K$ top-ranked utterances, for $K \in \{1, 5, 10\}$.

\vskip0.1cm \noindent{\bf Representational similarity to image space~}
Representational similarity analysis (RSA) is a way of evaluating how pairwise similarities between objects are correlated in two object representation spaces \cite{kriegeskorte2008representational}. Here we compare cosine similarities among encoded utterances versus cosine similarities among vector representations of images. Specifically, we create two pairwise $N\times N$ similarity matrices: (a) among encoded utterances from the validation data, and (b) among images corresponding to each utterance in (a). Note that since there are five descriptions per image, each image is replicated five times in matrix (b). We then take the upper triangulars of these matrices (excluding the diagonal) and compute Pearson's correlation coefficient between them.
The image features for this evaluation are obtained from the  final fully connected layer of VGG-16 \cite{simonyan2014very} pre-trained on Imagenet \cite{ILSVRCarxiv14} and consist of 4096 dimensions. %\todo{i would perhaps put this before explaining the similarity matrix}

\subsection{Settings}
We preprocess the audio by extracting 12-dimensional mel-frequency cepstral
coefficients (MFCC) plus log of the total energy. We use 25 milisecond windows, sampled every 10 miliseconds.
Audio2vec and SegMatch models are trained for a maximum of 15 epochs with Adam, with learning rate 0.0002, and gradient clipping at 2.0. SegMatch uses margin $\alpha=0.2$. The encoder GRU  has 5 layers of 512 units. The convolutional layer has 64 channels, size of 6 and stride 3. The hidden layer of the attention MLP is 512. The GRU of the Audio2vec decoder has 512 hidden units; the size of the output of the projections $\mathbf{B}$ and $\mathbf{E}$ in SegMatch is also 512 units. 
For SegMatch the size of the erased center portion of the utterance is 30 frames.
We apply early stopping and report all the results of each model after the epoch for which it scored best on recall@10. 
When applying SegMatch on human data, each mini-batch includes utterances spoken only by one speaker: this is in order to discourage the model from encoding speaker-specific features.

\section{Results}
\subsection{Synthetic speech}
\begin{table*}[htbp]
% SEE https://github.com/gchrupala/vgs/blob/master/analysis/results.ipynb
\centering
\begin{tabular}{lrrr} \toprule
               & Recall@10 (\%) & Median rank & RSA$_{\text{image}}$ \\ \midrule
\color{gray} VGS
               & \color{gray} 
                 27   
                      & \color{gray}  
                            6 
                                             & \color{gray} 
                                               0.4 \\ 
SegMatch       & \bf 10   & \bf   37                & \bf 0.5 \\ 
Audio2vec-U    &      5   &      105                &  0.0 \\ 
Audio2vec-C    &      2   &      647                & 0.0 \\ 
Mean MFCC      &      1   &    1,414                & 0.0 \\
Chance         &      0   &    3,955                & 0.0 \\ \bottomrule
\end{tabular}
\caption{Results on Synthetically Spoken COCO. The row labeled VGS is the visually supervised model from \citet{chrupala2017representations}.}
\label{tab:results-coco}
\end{table*}

%\todo{Double check the RSA result for VGS}
Table~\ref{tab:results-coco} shows the evaluation results on synthetic speech. Representations learned by Audio2vec and SegMatch are compared to the performance of random vectors, mean MFCC vectors, as well as visually supervised representations (VGS, model from \citet{chrupala2017representations}). Audio2vec works better than chance and mean MFCC on paraphrase retrieval, but does not correlate with the visual space. SegMatch works much better than Audio2vec according to both criteria. It does not come close to VGS on paraphrase retrieval, but it does correlate with the visual modality even better.
\subsection{Human speech}

\paragraph{Places} This dataset only features a single caption per image and thus we only evaluate according to RSA: with both SegMatch and Audio2vec we found the correlations to be zero. 

\paragraph{Flickr8K} Initial experiments with Flickr8K were similarly unsuccessful. Analysis of the learned SegMatch representations revealed that in spite of partitioning the data by speaker for training, speaker identity can be decoded from them. 

\paragraph{Enforcing speaker invariance}
We thus implemented a version of SegMatch where an auxiliary speaker classifier is connected to the encoder via a gradient reversal operator \cite{pmlr-v37-ganin15}. This architecture optimizes the main loss, while at the same time pushing the encoder to remove information about speaker identity from the representation it outputs. In preliminary experiments we saw that this addition was able to prevent speaker identity from being encoded in the representations during the first few epochs of training. Evaluating this speaker-invariant representation gave contradictory results, shown in Table~\ref{tab:results-flickr8k}: very good scores on paraphrase retrieval, but zero correlation with visual space. 

\begin{table*}[htbp]
\centering
\begin{tabular}{lrrr} \toprule
               & Recall@10 (\%) & Median rank & RSA$_{\text{image}}$ \\ \midrule
\color{gray} VGS
               & \color{gray} 
                 15   
                      & \color{gray}  
                            17 
                                             & \color{gray} 
                                               0.2 \\ 
SegMatch       &     12   &   17                  & 0.0 \\ 
Mean MFCC      &      0   &    711                & 0.0 \\\bottomrule
\end{tabular}
\caption{Results on Flickr8K. The row labeled VGS is the visually supervised model from \citet{chrupala2017representations}. }
\label{tab:results-flickr8k}
\end{table*}

Further analysis showed that there seems to be an artifact in the Flickr8K data where spoken captions belonging to consecutively numbered images share some characteristics, even though the images do not. As a side effect, this causes captions belonging to the same image to also share some features, independent of their semantic content, leading to high paraphrasing scores. The artifact may be due to changes in data collection procedure which affected some aspect of the captions in ways which correlate with their sequential ordering in the dataset.

If we treat the image ID number as a regression target, and the first two principal components of the SegMatch representation of one of its captions as the predictors, 
we can account for about 12\% of the holdout variance in IDs using a non-linear model (using either K-Nearest Neighbors or Random Forest). This effect disappears if we arbitrarily relabel images. 

\section{Conclusion}
For synthetic speech the SegMatch approach to inducing utterance embeddings shows very promising performance. Likewise, previous work has shown some success with word-segmented speech. There remain challenges in carrying over these results to natural, unsegmented speech. Word segmentation is a highly non-trivial research problem in itself and the variability of spoken language is a serious and intractable confounding factor. 

Even when controlling for speaker identity there are still superficial features of the speech signal which make it easy for the model to ignore the semantic content. Some of these may be due to artifacts in datasets and thus care is needed when evaluating unsupervised models of spoken language: for example use of multiple evaluation criteria may help spot spurious results. In spite of these challenges, in future we want to further explore the effectiveness of enforcing desired invariances via auxiliary classifiers with gradient reversal.

%\section{Acknowledgements}

\bibliographystyle{acl_natbib}
\bibliography{biblio}
\end{document}

%% file: abstract.tex
In the domain of unsupervised learning most work on speech has focused on discovering  low-level constructs such as phoneme inventories or word-like units. In contrast, for written language, where there is a large body of work on unsupervised induction of semantic representations of words, whole sentences and longer texts. In this study we examine the challenges of adapting these approaches from written to spoken language. We conjecture that unsupervised learning of the semantics of spoken language  becomes feasible if we abstract from the surface variability. We simulate this setting with a dataset of utterances spoken by a realistic but uniform synthetic voice. We evaluate two simple unsupervised models which, to varying degrees of success, learn semantic representations of speech fragments. Finally we present inconclusive results on human speech, and discuss the challenges inherent in learning distributional semantic representations on unrestricted natural spoken language.